\documentclass{article}

\usepackage{arxiv}

\usepackage[utf8]{inputenc} 
\usepackage[T1]{fontenc}    
\usepackage{hyperref}       
\usepackage{url}            
\usepackage{booktabs}       
\usepackage{amsfonts}       
\usepackage{nicefrac}       
\usepackage{microtype}      
\usepackage{lipsum}
\usepackage{graphicx}
\graphicspath{ {./images/} }
\usepackage{graphicx}
\usepackage[onehalfspacing]{setspace}

\usepackage{tocbasic}
\usepackage{booktabs}
\usepackage{multicol}
\usepackage{fontawesome5}
\usepackage{multirow}
\usepackage{algorithm}
\usepackage{algpseudocode}
\usepackage{caption}
\usepackage{tabularray}
\usepackage{subcaption}
\usepackage{array}

\title{Mixture of Modular Experts: Distilling Knowledge from a Multilingual Teacher into Specialized Modular Language Models}

\author{
  Mohammed Al-Maamari \\
  Chair of Data Science\\
  University of Passau\\
  Germany, Passau \\
  \texttt{Mohammed.Al-Maamari@uni-passau.de} \\
   \And
 Mehdi Ben Amor \\
  Chair of Data Science\\
  University of Passau\\
  Germany, Passau \\
  \texttt{Mehdi.BenAmor@uni-passau.de} \\
  \And
 Michael Granitzer \\
  Chair of Data Science\\
  University of Passau\\
  Germany, Passau \\
  \texttt{Michael.Granitzer@uni-passau.de} \\
}

\begin{document}
\maketitle
\begin{abstract}
This research explores the integration of Knowledge Distillation (KD) and Mixture of Experts (MoE) to create modular, efficient, and specialized multilingual language models. The primary objectives include evaluating adaptive versus fixed alpha methods in KD, developing and comparing modular MoE architectures in handling multi-domain inputs and preventing catastrophic forgetting.

We address the computational challenges of large language models (LLMs) and the need for modular models. KD compresses LLMs into smaller models for efficiency, while MoE architectures enhance modularity by combining multiple experts for specialized tasks. Experiments showed that adaptive and fixed alpha methods in KD yielded similar performance, with marginal improvements from adaptive alpha. The combined loss approach slightly outperformed alternating losses, providing more stable learning. The router, trained to classify input sequences into English, French, German, or Python, achieved high accuracy precision, recall, and F1 score of 99.95\%, with Logistic Regression as the most effective classifier.

Evaluation of modular MoE architectures revealed that the Pre-trained Language Experts (PLE) setup and Joint Expert Embedding Training (JEET) demonstrated similar performance, while the MoE with Common Expert (MoE-CE) setup showed slightly lower performance. However, when including a common expert in MoE-CE, its performance approaches that of both PLE and JEET. The study on catastrophic forgetting indicated that sequential training led to significant forgetting, while single-session training with balanced batches approach, and MoE approach mitigated this issue effectively. The MoE architecture preserved knowledge across multiple languages, demonstrating its effectiveness. We open-sourced the dataset \footnote{\url{https://zenodo.org/doi/10.5281/zenodo.12677631}}, the balanced dataset creation tool \footnote{\url{https://github.com/padas-lab-de/multi-language-dataset-creator}}, and the research codebase \footnote{\url{https://github.com/ModMaamari/mixture-modular-experts}}.

\end{abstract}


\section{Introduction}
Language models (LMs) are pivotal in Natural Language Processing (NLP), facilitating a variety of tasks such as machine translation \cite{lembersky2012language}, sentiment analysis \cite{zhang2023sentiment}, and text generation \cite{brown2020language}. Despite their potential, large-scale models encounter challenges like computational inefficiency, limited adaptability, and catastrophic forgetting. Our study explores the amalgamation of Knowledge Distillation (KD) and Mixture of Experts (MoE) to mitigate these challenges, aiming to improve efficiency, modularity, and specialization in language models.

Transformers, the backbone of many large models, require substantial computational resources \cite{ganesh2021compressing}, which hampers their scalability and accessibility. The increasing complexity and size associated with supporting more languages and domains adversely affect training durations and generalization abilities \cite{zhao2023survey}. Additionally, fine-tuning for specific tasks consumes significant resources and often falls short of achieving optimal outcomes \cite{hu2021lora}. Catastrophic forgetting is a major hurdle, particularly in models handling multiple languages and domains, as they tend to lose previously acquired knowledge when exposed to new data \cite{goodfellow2013empirical}.

Specialized models, when trained on narrow domains such as programming languages, have demonstrated superiority in specific tasks like code completion and bug detection over their general-purpose counterparts \cite{jiang2023impact}. Introducing modularity into neural network design enhances flexibility, scalability, and maintainability, enabling updates to individual network segments without necessitating a complete retraining.

This research primarily focuses on exploring various integration strategies of KD and MoE to create specialized, efficient, and modular language models. While we employed straightforward knowledge distillation techniques, reaching state-of-the-art knowledge distillation was not our objective. Instead, our primary goal was to investigate the feasibility of different integration methods of KD and MoE. KD is the process where smaller student models learn to mimic the behavior of a larger, more capable teacher model using their probabilistic outputs \cite{hinton2015distilling}. MoE architectures, on the other hand, dynamically delegate tasks to specialized models, thereby enhancing performance across varied domains and languages \cite{shazeer2017outrageously}.

Our research objectives include evaluating adaptive versus fixed alpha methods in KD, training a router to efficiently direct inputs to the appropriate experts, and comparing various MoE architectures to determine their effectiveness in handling multi-domain inputs and in averting catastrophic forgetting. This study contributes to the development of more adept and effective NLP systems that can support a broad spectrum of applications.

\section{Related Work}

\subsection{Knowledge Distillation}

"DistilBERT" by Sanh et al. \cite{sanh2019distilbert} introduces a method to distill BERT into a smaller, faster model while retaining most of its performance. By leveraging a triple loss function (language modeling, distillation, and cosine-distance losses), DistilBERT reduces model size by 40\% and maintains 97\% of BERT's language understanding capabilities. This approach makes the model suitable for deployment in environments with constrained computational resources. DistilBERT's training involves using every second layer from the teacher model to retain inductive biases and employs techniques like gradient accumulation and dynamic masking.

"MiniLLM" by Gu et al. \cite{gu2023minillm} proposes using reverse Kullback-Leibler divergence (KLD) for knowledge distillation to address the issue of overestimating low-probability regions in the teacher model's distribution. The method minimizes the difference between the teacher and student model distributions using a conditional text generation framework. Techniques such as single-step decomposition and teacher-mixed sampling are employed to stabilize and accelerate the training process. Our work adopts a simplified version of reverse KLD at the word level, focusing on efficiently capturing the teacher model's probabilistic characteristics.

\subsection{Mixture of Experts}

"Mixtral of Experts" by Jiang et al. \cite{jiang2024mixtral} presents Mixtral 8x7B, a Sparse Mixture of Experts (SMoE) language model. The model dynamically selects two out of eight feedforward blocks per token at each layer, optimizing computational resource usage. Mixtral's transformer model incorporates MoE layers with a routing mechanism to allocate tokens to experts. Evaluations show superior performance in various benchmarks, highlighting the model's efficiency and effectiveness.

"Branch-Train-MiX" by Sukhbaatar et al. \cite{sukhbaatar2024branch} investigates methods for training LLMs across multiple specialized domains. The Branch-Train-MiX (BTX) method involves branching from a seed model, training domain-specific experts, and integrating them into a unified MoE model. This approach improves training efficiency and model performance by leveraging parallelism and specialization. BTX outperforms baselines like Llama-2 in accuracy and computational efficiency.

"Branch-Train-Merge" by Li et al. \cite{li2022branch} introduces the Branch-Train-Merge (BTM) algorithm, which enhances the efficiency of training large language models. BTM facilitates independent training of subparts of the model on different data subsets, reducing communication overhead. The approach involves three steps: branching, training, and merging. BTM achieves improved perplexities and higher updates per second due to reduced communication overhead, making it a scalable and efficient training paradigm.

Our research integrates MoE with Knowledge Distillation (KD) to develop specialized multilingual models. Unlike Mixtral and BTX, which focus on token-level routing and parallel training of domain-specific experts, our work emphasizes sequence-level routing and the integration of KD with MoE. This approach aims to address multi-domain adaptability and reduce catastrophic forgetting, contributing to the development of modular and efficient language models.

\section{Methods}

This section describes the methodologies, tools, and algorithms used in this research, focusing on dataset preparation, model training, knowledge distillation techniques, and MoE architecture design.

\subsection{Dataset Preparation}

The dataset comprises multilingual text data in English, German, French, and Python code. The primary sources are the Wiki40B dataset \cite{guo2020wiki} for natural languages and the CodeParrot GitHub "codeparrot/github-code-clean" dataset \cite{codeparrot_github_code_clean} for Python code. The Wiki40B dataset includes 2,926,536 English, 1,227,206 French, and 1,554,908 German training samples, ensuring balanced token counts across these languages. Python data contains 645K Python code files, filtered and balanced based on code length to ensure comprehensive coverage.

\begin{table}[h]
    \centering
    \caption{Number of Articles per Language}
    \begin{tabular}{lcc}
        \toprule
        \textbf{Language} & \textbf{Training Samples} & \textbf{Validation Samples} \\
        \midrule
        English & 2,926,536 & 163,597 \\
        French & 1,227,206 & 68,655 \\
        German & 1,554,908 & 86,068 \\
        \bottomrule
    \end{tabular}
\end{table}

\subsection{Tokenization}

Byte Pair Encoding (BPE) \cite{gage1994new} was used for tokenization, with a vocabulary size of 32,000 tokens. The tokenizer was trained on a balanced dataset of the four languages, using special tokens such as \texttt{<unk>}, \texttt{<s>}, and \texttt{</s>}. This setup ensured efficient vocabulary usage and compatibility with the models.

\begin{figure}[h]
    \centering
    \includegraphics[width=\textwidth]{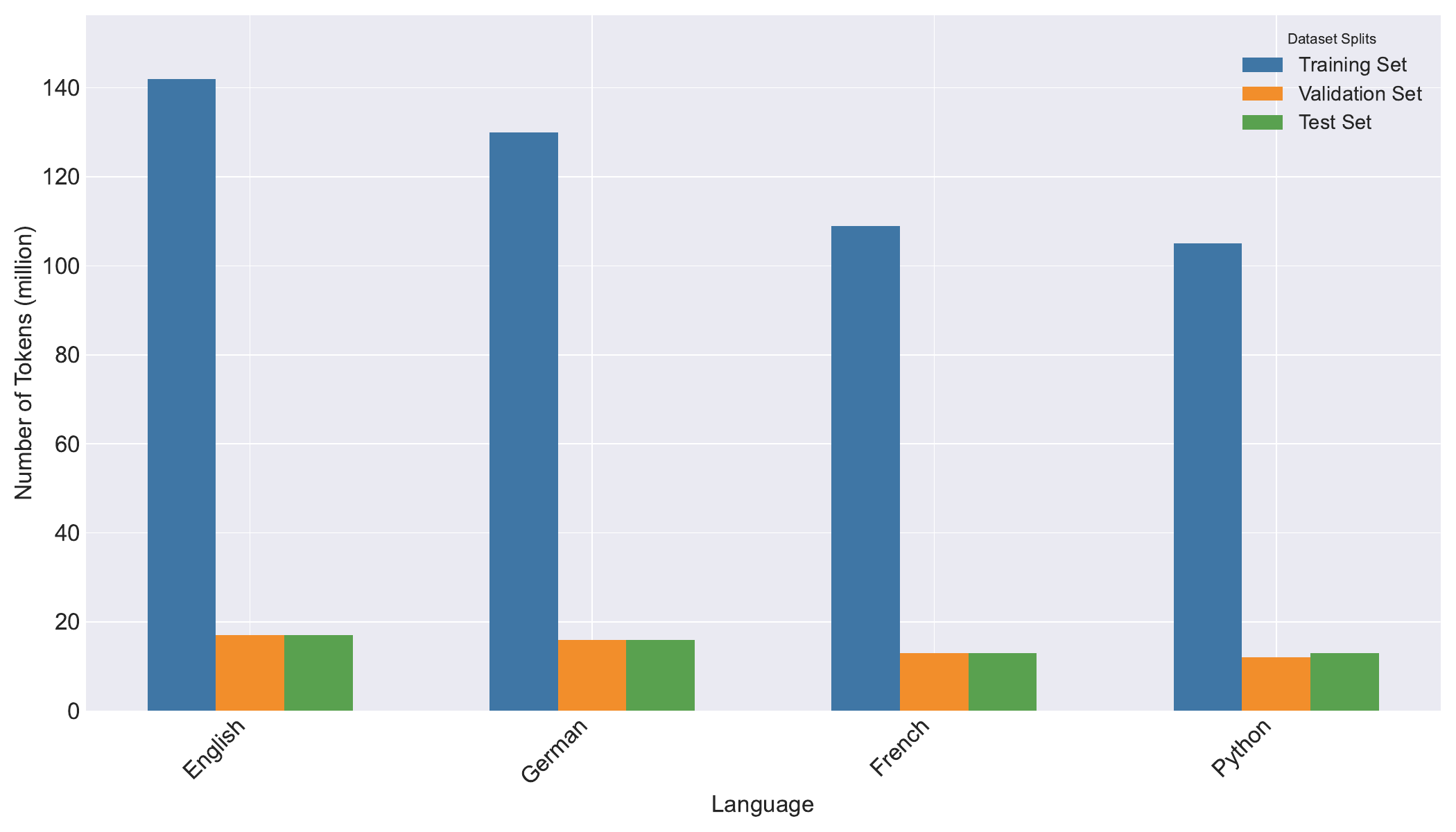}
    \caption{Dataset Splits for Each Language}
    \label{fig:data-splits}
\end{figure}

\subsection{Teacher Model Training}

The teacher model, a GPT-2 Medium with 340 million parameters, was trained on the multilingual dataset. Initial attempts with Mistral 1.6B and Phi 1.34B configurations faced stability issues, leading to the selection of GPT-2 Medium for its balance of performance and computational efficiency. The training process involved a context length of 1024 tokens, a virtual batch size of 512, and optimization techniques such as gradient accumulation and clipping. Training was conducted on 2 A100 GPUs using DeepSpeed and Accelerate libraries.

\begin{figure}[h]
    \centering
    \includegraphics[width=\textwidth]{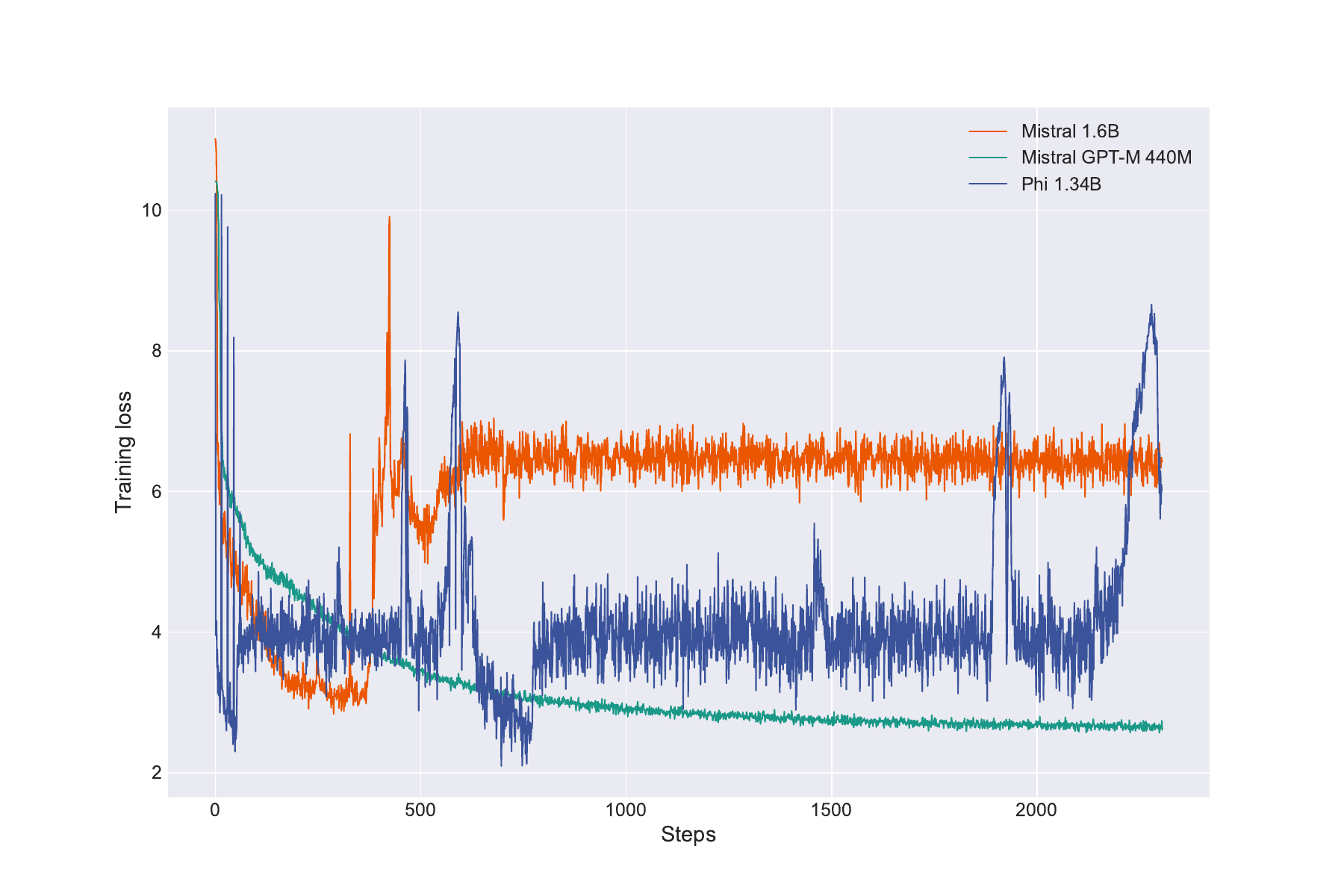}
    \caption{Learning Curve for Mistral 1.6B, Mistral GPT-M 440M, Phi 1.34B}
    \label{fig:learning-curves}
\end{figure}

\subsection{Knowledge Distillation}

Knowledge distillation (KD) involved transferring knowledge from the GPT-2 Medium teacher model to smaller student models. The student models were trained to replicate the teacher's outputs using a combined loss function incorporating Cross-Entropy Loss and Reverse Kullback-Leibler (RKL) Divergence Loss \cite{gu2023minillm}. The adaptive alpha method dynamically adjusted the weights of these losses based on training progress.

\begin{figure}[h]
    \centering
    \includegraphics[width=\textwidth]{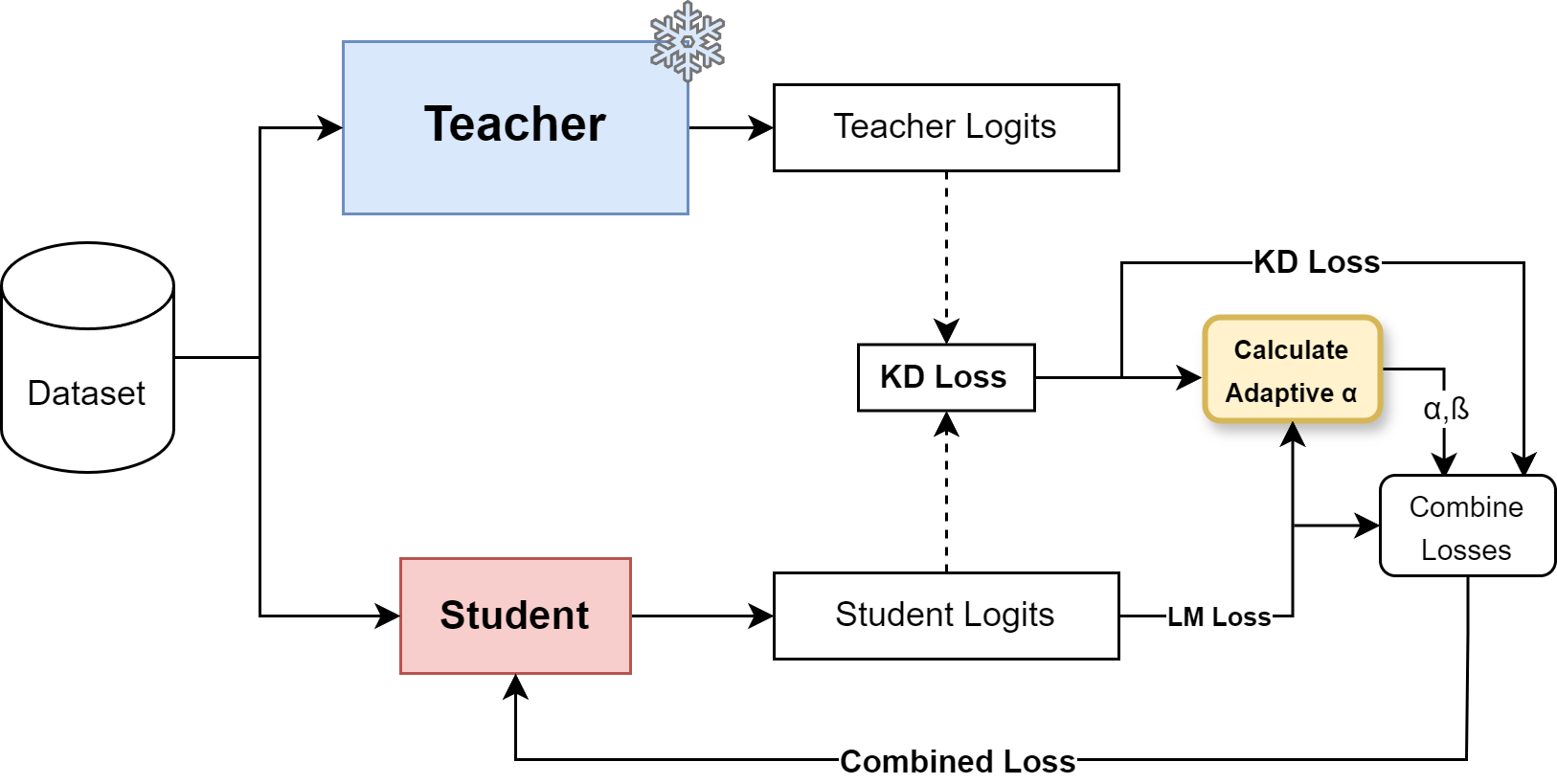}
    \caption{Our Knowledge Distillation Process}
    \label{fig:our-kd-process}
\end{figure}

\begin{equation}
    \mathcal{L}_{total} = \alpha \cdot \mathcal{L}_{LM} + \beta \cdot \mathcal{L}_{KD}
    \label{eq:total_loss}
\end{equation}

\subsection{Mixture of Experts Architecture}

The Mixture of Experts (MoE) architecture includes multiple specialized sub-models (experts) and a router mechanism. Three setups were evaluated:

\textbf{Setup 1: Pre-trained Language Experts (PLE)} - Experts pre-trained independently via KD from the teacher model, handling specific languages.

\textbf{Setup 2: Joint Expert Embedding Training (JEET)} - Experts trained concurrently with a shared embedding layer, maintaining modularity.

\textbf{Setup 3: MoE with Common Expert (MoE-CE)} - Includes a common expert trained on all languages, sharing the embedding layer with specialized experts.

\begin{figure}[h]
    \centering
    \includegraphics[width=0.8\textwidth]{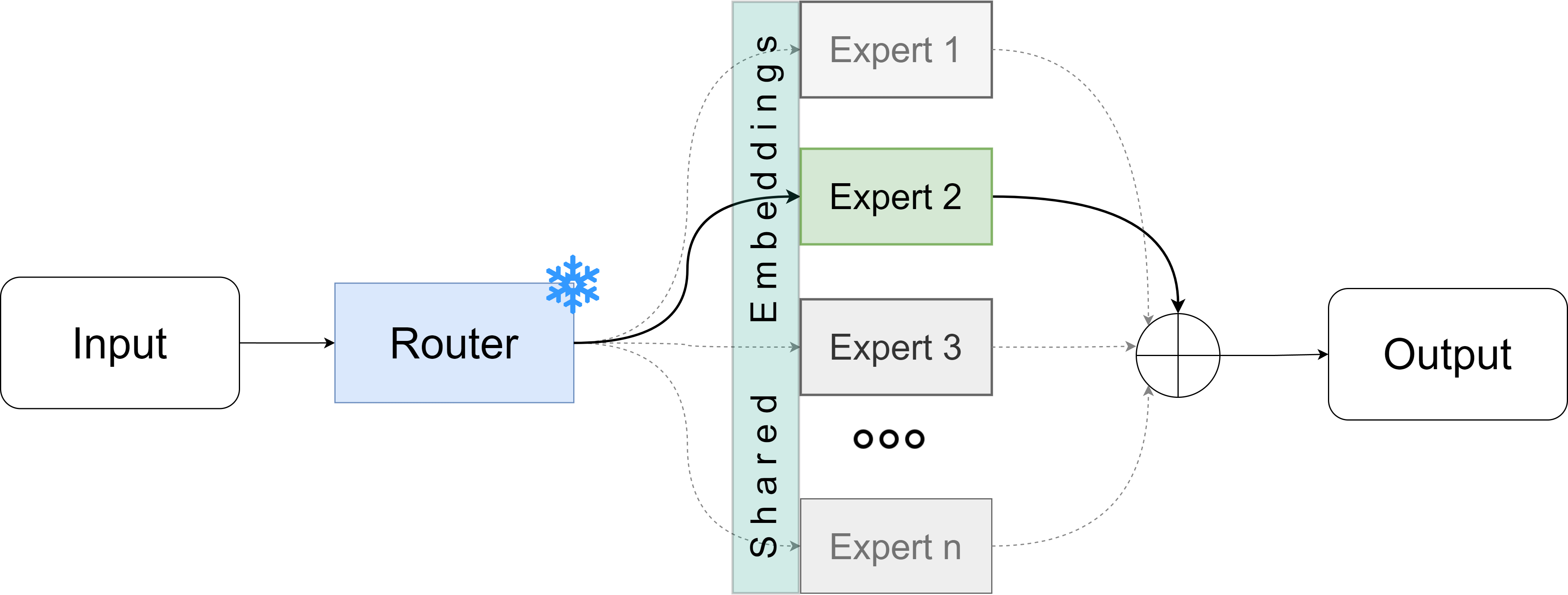}
    \caption{Architecture of the Joint Expert Embedding Training MoE Setup}
    \label{fig:moe_jeet}
\end{figure}

\subsection{Router}

The router, pre-trained to classify inputs into English, French, German, or Python, achieved high accuracy (99.95\%) using TF-IDF vectorization and Logistic Regression. It dynamically selects the appropriate expert during inference, optimizing the allocation of computational resources.

\subsection{Training and Inference}

During training, data was batched by language, and the router directed each batch to the corresponding expert. This approach ensured specialization without interference. During inference, the system handled mixed-language batches, maintaining efficiency and specialization.

\begin{figure}[h]
    \centering
    \includegraphics[width=0.8\textwidth]{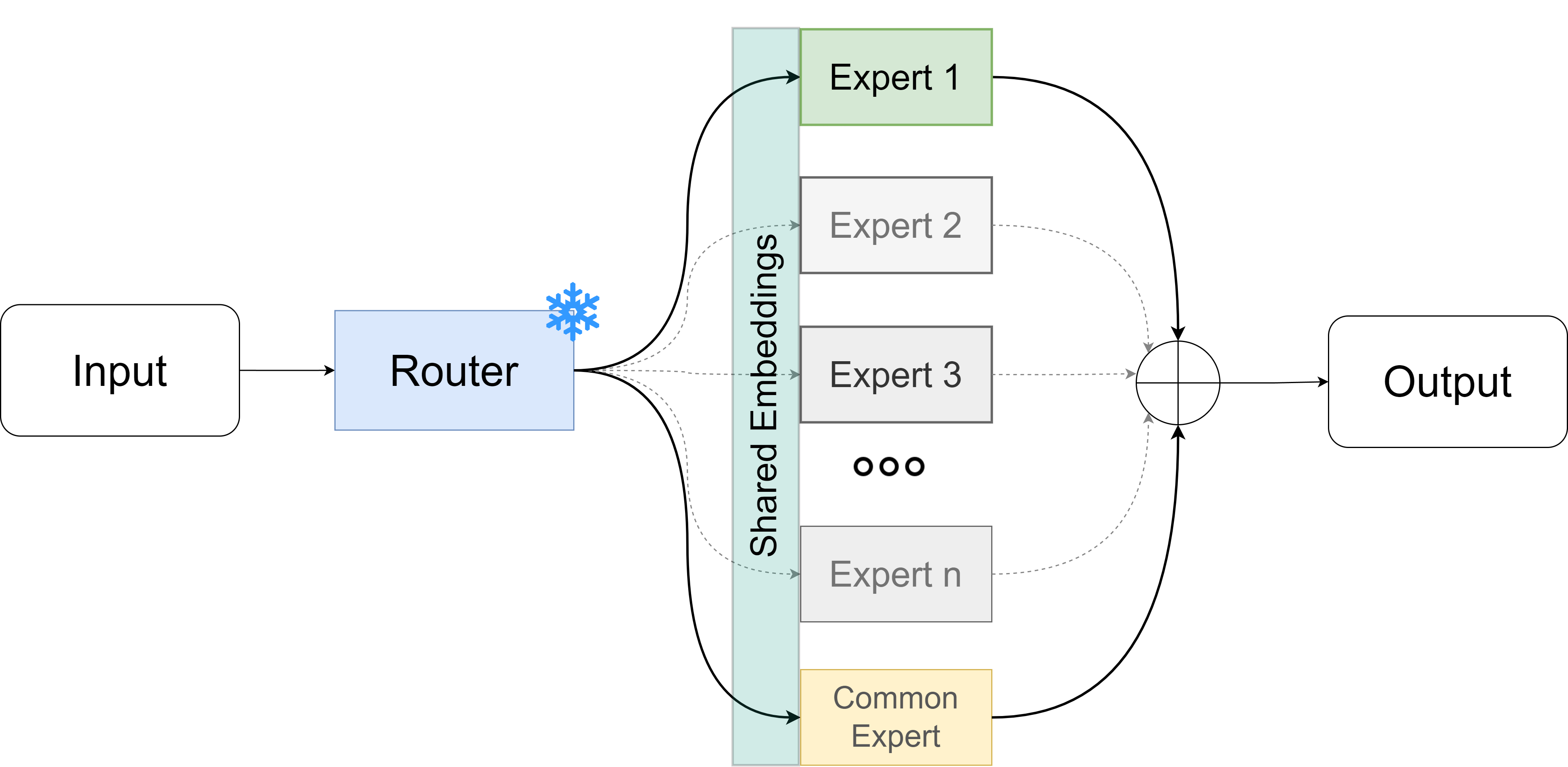}
    \caption{Architecture of the MoE with Common Expert Setup}
    \label{fig:moe_ce}
\end{figure}

\begin{table}[h]
    \centering
    \caption{Model Configurations and Hyperparameters}
    \label{tab:model-configurations}
    \begin{tabular}{lccr}
        \hline
        \textbf{Model} & \textbf{Parameters} & \textbf{Configuration Source} & \textbf{Hyperparameters} \\
        \hline
        Mistral 1.6B & 1.6B & Phi1.5 & Custom \\
        Phi 1.34B & 1.34B & Phi1.5 & Standard \\
        GPT-2 110M & 110M & GPT-2 & Standard \\
        GPT-2 340M & 340M & GPT-2 & Standard \\
        Distil-GPT2 & 82M & Distil-GPT2 & Standard \\
        \hline
    \end{tabular}
\end{table}

\begin{figure}[h]
    \centering
    \includegraphics[width=0.8\textwidth]{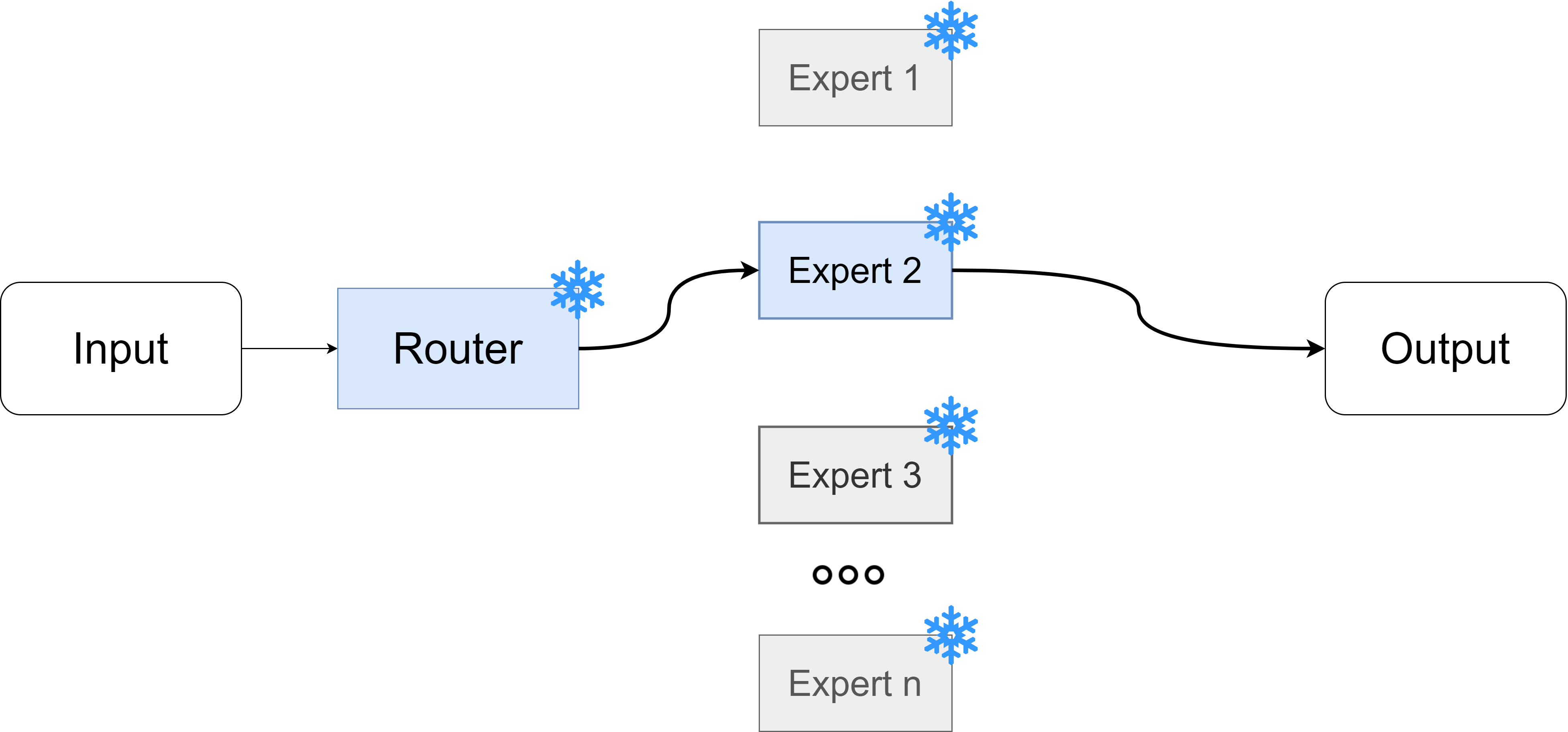}
    \caption{Architecture of the Pre-trained Language Experts}
    \label{fig:moe_frozen}
\end{figure}

\section{Results}

The experiments focused on evaluating different aspects of the Mixture of Experts (MoE) architecture, including Knowledge Distillation (KD) methods, router training, and the impact of modular MoE architectures on system performance.

\subsection{Experimental Setup}

The experimental setup involved two distinct KD approaches: independent KD and sequential KD. Independent KD trained separate student models for each language (English, French, German, and Python) using the outputs of the teacher model. Sequential KD distilled a single student model sequentially with knowledge from each language. The evaluation metrics used were perplexity and cross-entropy loss. The training hyperparameters, including a context length of 1024 tokens and a vocabulary size of 32,000, were consistent with those used for the teacher model. The MoE training utilized the same tokenizer and frameworks such as DeepSpeed and Accelerate to manage computational demands.

\subsection{Adaptive vs. Fixed Alpha}
\label{sec:results-adaptive_vs_fixed_alpha}

\textbf{Research Question 1 (RQ1):} \textit{What is the effectiveness of adaptive vs. fixed alpha methods in Knowledge Distillation?}

\begin{figure}[h]
    \centering
    \includegraphics[width=0.8\textwidth]{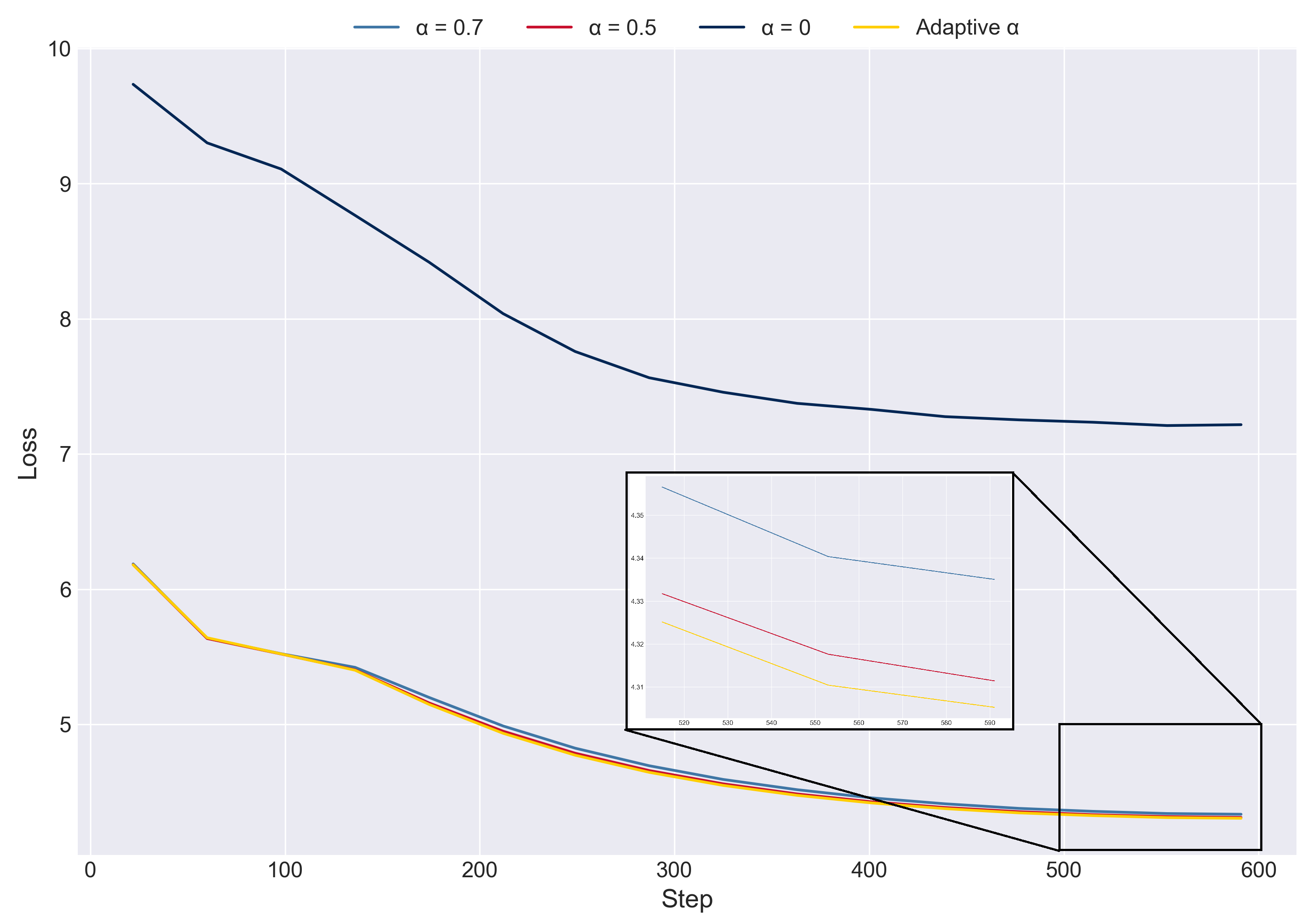}
    \caption{Evaluation loss for Adaptive Alpha and different values of Fixed Alpha}
    \label{fig:adaptive_vs_fixed_alpha}
\end{figure}

The adaptive alpha method, which dynamically adjusts weights during training, was hypothesized to perform better than the fixed alpha method. The results, shown in \autoref{fig:adaptive_vs_fixed_alpha}, indicated that the adaptive alpha method outperformed the fixed alpha method by a small margin (0.01 improvement in evaluation loss). Testing different fixed alpha values showed that an alpha of 0.5 performed nearly as well as the adaptive method, suggesting minimal benefits of dynamic adjustment in this setup.

\subsection{Alternating Losses (AL) VS Combined Losses (CL)}
\label{sec:results-alternating_vs_combined_losses}

\textbf{Research Question 2 (RQ2):} \textit{What is the impact of alternating losses during training on model convergence and performance?}

\begin{figure}[h]
    \centering
    \includegraphics[width=0.8\textwidth]{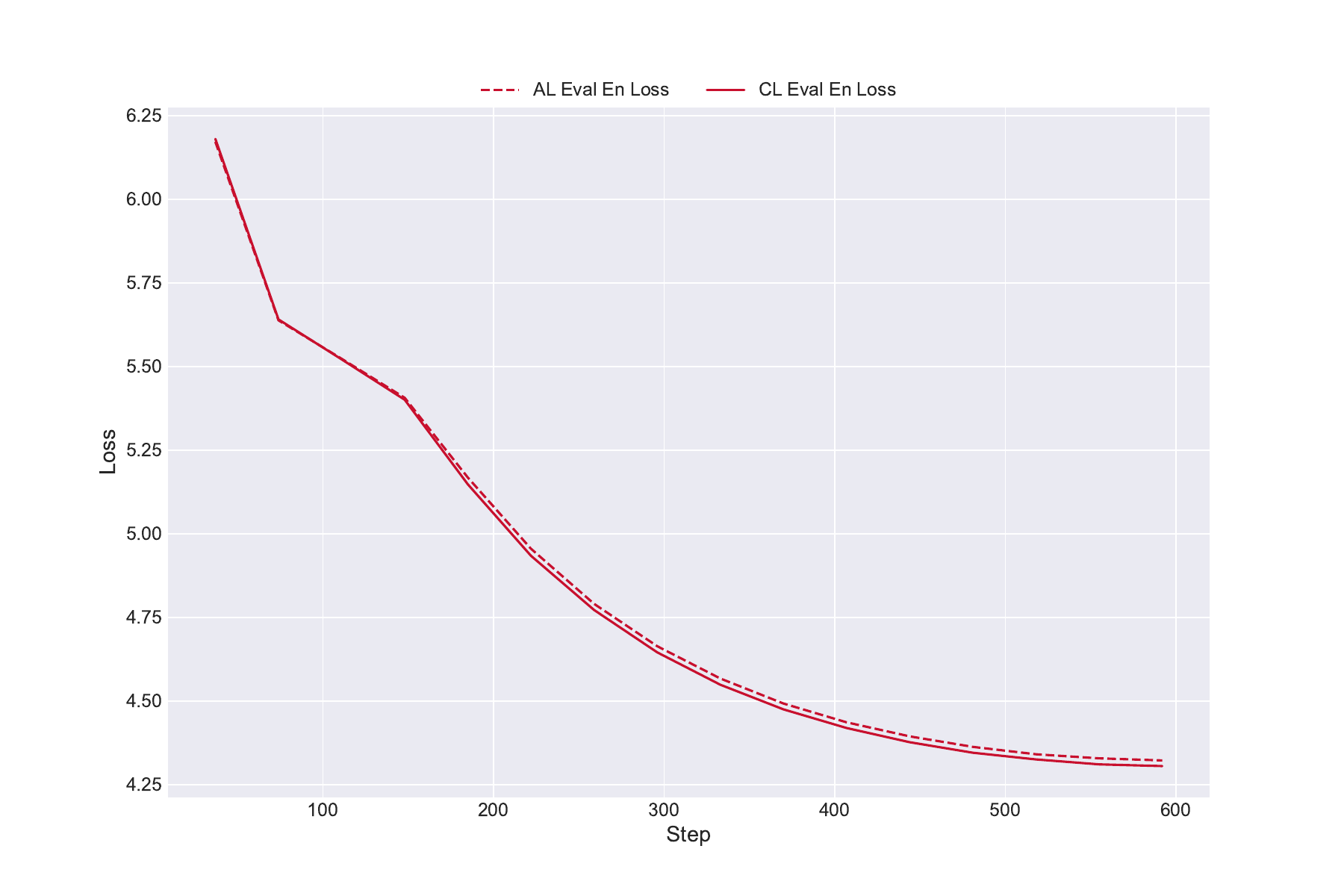}
    \caption{Evaluation Loss for Combined Loss vs. Alternating Losses Methods}
    \label{fig:combined_vs_alternating_losses}
\end{figure}

The hypothesis was that alternating KD\_loss and LM\_loss would enhance model performance. However, the results, shown in \autoref{fig:combined_vs_alternating_losses}, indicated that the combined loss method performed slightly better, with an evaluation loss marginally lower than the alternating losses method (4.305 vs. 4.322). This suggests that consistently applying both losses at each training step provides a more effective learning signal.

\subsection{The Router}
\label{sec:results-router}

\textbf{Research Question 3 (RQ3):} \textit{Can we train a router that accurately routes input sequences to one of the four experts, classifying the input sequence into one of the four classes [En, Fr, De, Py]?}

\begin{figure}[h]
    \centering
    \includegraphics[width=0.8\textwidth]{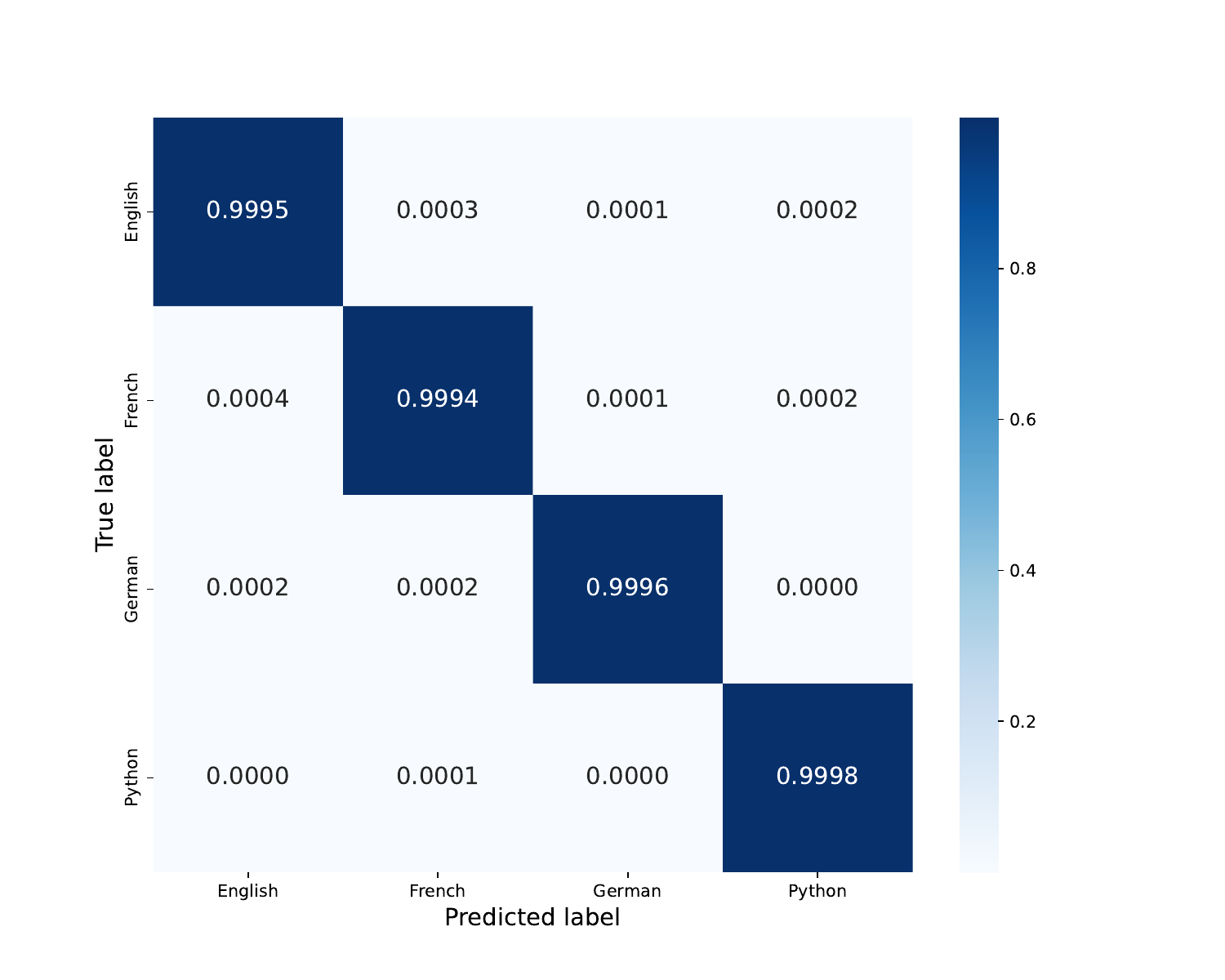}
    \caption{Confusion Matrix for the Router}
    \label{fig:router_confusion_matrix}
\end{figure}

Various classifiers were evaluated, with the Logistic Regression classifier achieving the highest performance (99.95\% accuracy). The confusion matrix in \autoref{fig:router_confusion_matrix} and \autoref{tab:router_performance_metrics} confirmed the router's effectiveness in accurately classifying input sequences.

\begin{table}[h!]
    \centering
    \caption{Performance Metrics of Different Classifiers for Router Training}
    \label{tab:router_performance_metrics}
    \begin{tabular}{lcccc}
        \toprule
        \textbf{Classifier} & \textbf{Accuracy $\uparrow$
} & \textbf{Precision $\uparrow$
} & \textbf{Recall $\uparrow$
} & \textbf{F1 Score $\uparrow$
} \\
        \midrule
        Logistic Regression & 0.9995 & 0.9995 & 0.9995 & 0.9995 \\
        SGD Classifier & 0.9993 & 0.9993 & 0.9993 & 0.9993 \\
        Random Forest & 0.9995 & 0.9995 & 0.9995 & 0.9995 \\
        \bottomrule
    \end{tabular}
\end{table}

\subsection{Modular MoE Model Design}
\label{sec:results-moe_model_design}

\textbf{Research Question 4 (RQ4):} \textit{How do the three MoE architectures compare in terms of performance and robustness?}

Three MoE setups were compared: Pre-trained Language Experts (PLE), Joint Expert Embedding Training (JEET), and MoE with Common Expert (MoE-CE). The results, summarized in \autoref{tab:moe_architecture_comparison}, showed that PLE achieved the best perplexities for English and German, while JEET performed best for French and Python. MoE-CE, when using the common expert, showed performance approaching that of PLE and JEET, highlighting the benefits of including a common expert.

\begin{table}[h]
    \centering
    \caption{Perplexity Metrics for Different MoE Architectures}
    \label{tab:moe_architecture_comparison}
    \begin{tabular}{lcccc}
        \toprule
        \textbf{Architecture} & \textbf{English $\downarrow$} & \textbf{French $\downarrow$} & \textbf{German $\downarrow$} & \textbf{Python $\downarrow$} \\
        \midrule
        PLE & \textbf{74.09} & 20.30 & \textbf{39.86} & 28.92 \\
        JEET & 75.79 & \textbf{20.12} & 40.38 & \textbf{27.02} \\
        MoE-CE w/o CE & 90.83 & 23.24 & 47.75 & 29.89 \\
        MoE-CE + CE & 78.96 & 20.91 & 41.92 & 27.16 \\
        \bottomrule
    \end{tabular}
\end{table}

\subsection{Impact of Adding a Common Expert to the MoE System}
\label{sec:results-impact_of_common_expert}

\textbf{Research Question 5 (RQ5):} \textit{Does adding a common expert improve the overall performance of the MoE system and the performance of each expert independently?}

The addition of a common expert in MoE-CE generally improved performance, as shown in \autoref{tab:moe_ce_settings_comparison}. The overall perplexity was reduced by 5.69 points with the inclusion of the common expert, confirming its positive impact.

\begin{table}[h!]
    \centering
    \caption{Perplexity scores for Different Inference Settings in MoE-CE}
    \label{tab:moe_ce_settings_comparison}
    \begin{tblr}{ccccccc}
        \hline
        \textbf{CommonExpert} & \textbf{RoutableExperts} &  \textbf{En $\downarrow$} & \textbf{Fr $\downarrow$} & \textbf{De $\downarrow$} & \textbf{Py $\downarrow$} & \textbf{All $\downarrow$} \\
        \hline
        \faCheck & \textbf{En, Fr, De, Py} & \textbf{78.96} & \textbf{20.91} & \textbf{41.92} & \textbf{27.16} & \textbf{42.24} \\
        \faTimes & En, Fr, De, Py & 90.83 & 23.24 & 47.75 & 29.89 & 47.93 \\
        \hline
        \faCheck & En, Py & \underline{78.96} & 36.55 & 79.95 & \underline{27.16} & 55.65 \\
        \faTimes & En, Py & \underline{90.83} & 116.04 & 325.28 & \underline{29.89} & 140.51 \\
        \hline[dashed]
        \faCheck & Fr, Py & 139.09 & \underline{20.91} & 77.64 & \underline{27.16} & 66.20 \\
        \faTimes & Fr, Py & 534.64 & \underline{23.24} & 305.62 & \underline{29.89} & 223.35 \\
        \hline[dashed]
        \faCheck & De, Py & 140.33 & 37.10 & \underline{41.92} & \underline{27.16} & 61.63 \\
        \faTimes & De, Py & 532.25 & 124.18 & \underline{47.75} & \underline{29.89} & 183.52 \\
        \hline
        \faCheck & En & \underline{78.96} & 36.55 & 79.95 & 53.65 & 62.28 \\
        \faTimes & En & \underline{90.83} & 116.04 & 325.28 & 203.50 & 183.91 \\
        \hline[dashed]
        \faCheck & Fr & 139.09 & \underline{20.91} & 77.64 & 68.62 & 76.56 \\
        \faTimes & Fr & 534.64 & \underline{23.24} & 305.62 & 577.22 & 360.18 \\
        \hline[dashed]
        \faCheck & De & 140.33 & 37.10 &  \underline{41.92} & 55.11 & 68.62 \\
        \faTimes & De & 532.25 & 124.18 & \underline{47.75} & 240.68 & 236.22 \\
        \hline[dashed]
        \faCheck & Py & 113.47 & 38.18 & 80.44 &   \underline{27.16} & 64.81 \\
        \faTimes & Py & 284.49 & 141.93 & 379.81 & \underline{29.89} & 209.03 \\
        \hline[dashed]
        \faCheck & None & 104.01 & 27.66 & 57.72 & 38.03 & 56.86 \\
        \hline
    \end{tblr}
\end{table}

\subsection{Catastrophic Forgetting in Modular MoE Architecture VS Non-Modular Approaches}
\label{sec:results-catastrophic_forgetting}

\textbf{Research Question 6 (RQ6):} \textit{How does the modular MoE architecture compare to non-modular approaches in terms of catastrophic forgetting?}

Three experiments were conducted to compare the impact of catastrophic forgetting:
1. Sequentially distilling knowledge into a single student model.
2. Distilling knowledge into a single student model in one session.
3. Employing the MoE architecture with four separate student models.

\begin{figure}[h]
    \centering
    \includegraphics[width=0.8\textwidth]{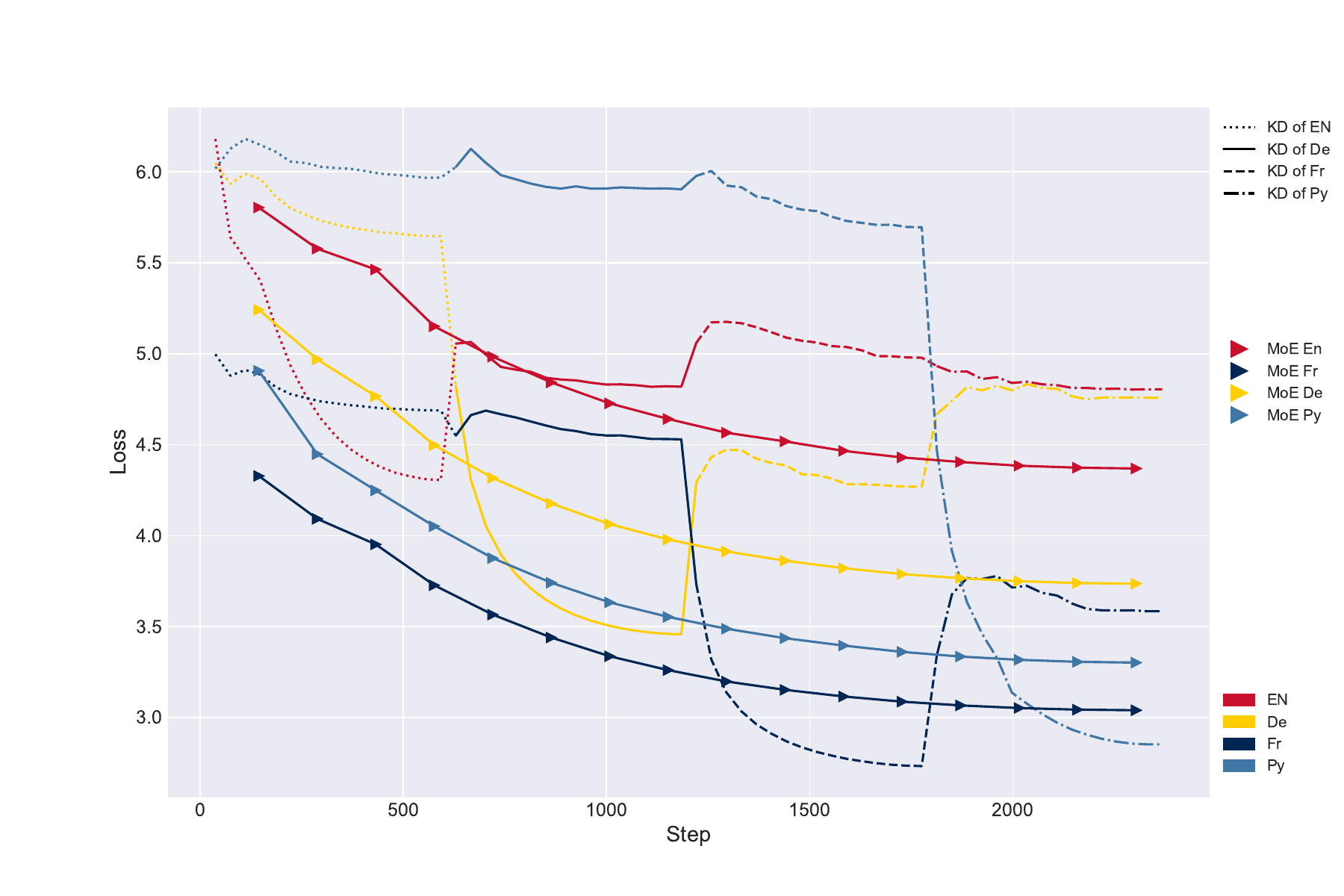}
    \caption{Evaluation Loss Comparison for Sequential Training (Experiment A) and MoE Architecture (Experiment C)}
    \label{fig:forgetting_comparison}
\end{figure}

The results, shown in \autoref{tab:forgotten_knowledge}, indicated significant catastrophic forgetting in the sequential setup (up to 38\% in German), whereas both single session distillation and MoE showed no catastrophic forgetting. The evaluation loss comparison for sequential training and MoE architecture is shown in \autoref{fig:forgetting_comparison}.

\begin{table}[h!]
    \centering
    \caption{Forgotten Knowledge in Different Experiments}
    \label{tab:forgotten_knowledge}
    \begin{tabular}{lcc}
        \toprule
        \textbf{Experiment} & \textbf{Language} & \textbf{Forgotten Knowledge $\downarrow$} \\
        \midrule
        \multirow{4}{*}{A (Sequential)} & English & 0.499 (12.0\%) \\
                                        & French & 0.851 (31.0\%) \\
                                        & German & 1.301 (38.0\%) \\
                                        & Python & N/A \\
        \midrule
        \multirow{4}{*}{B (Single Session)} & English & 0 (0\%) \\
                                            & French & 0 (0\%) \\
                                            & German & 0 (0\%) \\
                                            & Python & 0 (0\%) \\
        \midrule
        \multirow{4}{*}{C (MoE)} & English & 0 (0\%) \\
                                 & French & 0 (0\%) \\
                                 & German & 0 (0\%) \\
                                 & Python & 0 (0\%) \\
        \bottomrule
    \end{tabular}
\end{table}

The results confirmed that the modular MoE architecture effectively mitigates catastrophic forgetting, supporting the hypothesis that modularity allows for the retention of knowledge across multiple languages.

\section{Discussion}
\label{sec:discussion}

This section interprets and contextualizes the findings from our experiments, providing insights into the efficacy of different Knowledge Distillation (KD) methods and Mixture of Experts (MoE) architectures. It addresses the research questions, compares our results with existing literature, and reflects on the challenges and limitations encountered during the research.

\subsection{Interpretation of Findings}

\subsubsection{Adaptive vs. Fixed Alpha}
The comparison between adaptive alpha and fixed alpha methods, discussed in \autoref{sec:results-adaptive_vs_fixed_alpha}, revealed that both approaches yielded similar performance levels. The adaptive alpha approach showed a marginally better performance, but this advantage was minimal due to the consistency of the dataset used for training both the teacher and student models. The similarity in KD loss and cross-entropy loss changes likely contributed to the comparable performance of both methods.

\subsubsection{Alternating Losses vs. Combined Losses}
The experiment comparing alternating losses (AL) and combined losses (CL), as described in \autoref{sec:results-alternating_vs_combined_losses}, indicated that the combined loss approach slightly outperformed the alternating loss approach. The minimal difference suggests that the simplicity of the combination method—weighted averaging of the losses—may have masked potential benefits of alternating losses. More sophisticated methods of loss alternation may yield more pronounced differences and warrant further exploration.

\subsubsection{Router Performance}
The router's performance, detailed in \autoref{sec:results-router}, achieved high accuracy, precision, recall, and F1 scores. The distinctiveness of the four classes (English, German, French, and Python) and the balanced dataset used for training contributed to this success. The results confirm the router's capability to accurately classify input sequences and support the MoE architecture's performance.

\subsubsection{Modular MoE Model Design}
The performance comparison of the three MoE architectures, discussed in \autoref{sec:results-moe_model_design}, highlighted the strengths and limitations of each setup. Pre-trained Language Experts (PLE) and Joint Expert Embedding Training (JEET) performed comparably, with PLE excelling in English and German, and JEET in French and Python. MoE with Common Expert (MoE-CE) improved significantly with the inclusion of a common expert during inference, suggesting that a shared knowledge base can enhance performance in multi-language tasks.

\subsubsection{Catastrophic Forgetting in Modular vs. Non-Modular Approaches}
The study on catastrophic forgetting, as outlined in \autoref{sec:results-catastrophic_forgetting}, demonstrated that sequential training led to significant forgetting of previously learned languages, while single-session training and MoE approaches effectively mitigated this issue. The MoE approach, which assigns dedicated experts to each language, completely eliminated catastrophic forgetting, highlighting the effectiveness of modular architectures in preserving knowledge across multiple domains.

\subsection{Comparison with Existing Literature}

Our use of reverse Kullback-Leibler divergence (RKL) for knowledge distillation aligns with the approach presented in "MiniLLM: Knowledge Distillation of Large Language Models" \cite{gu2023minillm}, which demonstrated improved alignment between student and teacher models. Our findings validate the efficacy of RKL in our multilingual and modular model context.

In comparison to "Mixtral of Experts" \cite{jiang2024mixtral}, which uses a sparse MoE architecture, our approach emphasizes modularity and specialization. Our method allows for the utilization of any subset of experts without compromising their individual performance and includes a common expert to enhance performance across languages.

Our research also contrasts with "Branch-Train-MiX: Mixing Expert LLMs into a Mixture-of-Experts LLM" \cite{sukhbaatar2024branch}, which focuses on parallel training of domain-specific experts. Our approach combines KD and MoE, resulting in smaller, efficient student experts that maintain high performance. This dual focus on modularity and efficiency distinguishes our research.

\subsection{Challenges and Limitations}

\subsubsection{Challenges}
One significant challenge was the computational resource constraints, which necessitated the use of techniques such as gradient accumulation and efficient memory management. Ensuring balanced representation of the dataset also posed a challenge, requiring meticulous preprocessing and cleaning.

\subsubsection{Limitations}
The primary limitation is the scale of the dataset. Our training dataset size of 490 million tokens is considerably smaller than those used in training state-of-the-art language models. This limitation affects the generalizability of our findings to larger-scale applications. Additionally, the focus on a limited number of languages and a single programming language (Python) restricts the applicability of our findings.

\subsubsection{Future Work}
Future work should focus on addressing these limitations and exploring the scalability of our approach to larger datasets and more diverse languages and domains. Further research on the adaptive alpha method could help refine its implementation and identify scenarios where its benefits are more pronounced. Optimizing the training and integration process for the MoE architecture and investigating its applicability to other languages and domains will also be crucial.

\section{Conclusion}
\label{sec:conclusion}

\subsection{Summary of Contributions}
This research integrates Knowledge Distillation (KD) and Mixture of Experts (MoE) to develop modular, efficient, and specialized multilingual language models. The primary objectives were to evaluate adaptive versus fixed alpha methods in KD, compare modular MoE architectures, and address catastrophic forgetting.

\subsubsection{Knowledge Distillation}
The experiments comparing adaptive and fixed alpha methods in KD revealed similar performance, with the adaptive alpha method providing a slight improvement. The combined loss approach offered more stable learning dynamics compared to alternating losses.

\subsubsection{Mixture of Experts}
Three MoE architectures were assessed: Pre-trained Language Experts (PLE), Joint Expert Embedding Training (JEET), and MoE with Common Expert (MoE-CE). PLE and JEET performed similarly, while MoE-CE, without utilizing the common expert, lagged behind but demonstrated enhanced results with the inclusion of a common expert. This indicates the effectiveness of shared knowledge in improving performance across multiple languages.

\subsubsection{Router Performance}
The router, employing Logistic Regression for classification, achieved high accuracy and reliability in selecting the appropriate expert model for inputs, with accuracy, recall, precision, and F1-score all at 99.95\%.

\subsubsection{Catastrophic Forgetting}
Sequential training resulted in significant catastrophic forgetting, whereas single-session training and the MoE approach effectively mitigated this issue. The modular MoE architecture preserved knowledge across multiple languages, preventing catastrophic forgetting.

\subsection{Implications and Impact}
The integration of KD with MoE facilitates the development of modular, specialized, and efficient models that perform well across diverse tasks. The modularity of the MoE architecture enhances flexibility, allowing for the addition of new experts without retraining the entire system and addressing catastrophic forgetting by enabling the model to retain knowledge across multiple languages and domains.

\subsection{Challenges and Limitations}
The research was constrained by computational resources and a relatively small dataset of 490 million tokens, limiting the generalizability of the findings. Additionally, the focus on a limited number of languages and a single programming language indicates that further experimentation is needed to extend the approach to other languages, domains, and modalities.

\subsection{Future Work}
Future research should aim to scale the approach to larger datasets and more diverse languages and domains. Optimizing the training and integration process for the MoE architecture and exploring the applicability of the methods to other contexts are recommended. Further investigation into adaptive alpha methods and advanced loss functions for the common expert could provide deeper insights and enhance model performance.

\bibliographystyle{unsrt}  
\bibliography{main}  


\end{document}